\documentclass[letterpaper,10pt,conference]{ieeeconf}
\IEEEoverridecommandlockouts
\usepackage{amsmath}
\usepackage{amssymb}
\usepackage{booktabs}
\usepackage{graphicx}
\usepackage{tabularx}
\usepackage{cite}
\usepackage{url}
\usepackage{balance}
\usepackage{capt-of}
\usepackage[hidelinks]{hyperref}

\newcommand{\SEtwo}{\mathrm{SE}(2)}
\newcommand{\SEthree}{\mathrm{SE}(3)}
\newcommand{\SOthree}{\mathrm{SO}(3)}
\newcommand{\R}{\mathbb{R}}

\title{VOMMI: Collecting and Leveraging Portable Demonstrations for Mobile Manipulation}
\author{
\authorblockN{Yutian Zhang$^{1,*}$, Xingrui Xiong$^{4,*}$, Siyuan Ma$^{5,*}$, Donghao Wang$^{1,*}$, Yang Li$^2$, Jiawen Wen$^7$, Jiaqi Zhai$^3$, Liwen Yang$^1$, \\ Ce Hao$^6$, Haozhen Chi$^8$, Yangkun Zhu$^2$, Yifan Zhu$^2$, Xiaowen CHU$^7$, Dong Wei$^{6,\dagger}$, Qiaojun Yu$^{2,\dagger}$, Dibo Hou$^{1,\dagger}$}
\authorblockA{$^1$\textbf{ZJU} \quad $^2$\textbf{Shanghai AI Lab} \quad $^3$\textbf{DeepRobotics} \quad $^4$\textbf{Yale} \quad $^5$\textbf{THU} \quad $^6$\textbf{ZGCA} \quad $^7$\textbf{HKUST(GZ)} \quad $^8$\textbf{ZUST}\\ Equal contribution$^{*}$. Corresponding authors$^\dagger$}
}

\begin{document}
\maketitle
\thispagestyle{empty}
\pagestyle{empty}

\begin{abstract}
Portable mobile-manipulation demonstrations can help alleviate data scarcity for embodied intelligence, but obtaining reliable, low-cost, and robot-free motion supervision from RGB observations remains challenging.
Existing approaches often rely on teleoperation or specialized devices equipped with additional sensing hardware, while directly using estimated visual odometry (VO) trajectories can introduce inconsistencies due to accumulated drift and imperfect motion supervision.
We present the Visual-Odometry-Conditioned Mobile Manipulation Interface (VOMMI), a portable demonstration collection and learning framework that connects portable RGB demonstrations to vision-language-action (VLA) post-training through offline trajectory reconstruction and online visual-motion conditioning.    
VOMMI synchronizes body and hand views to capture navigation context and local object interactions without requiring human-robot kinematic correspondence calibration.
R2-VO refines offline demonstration trajectories using sparse geometric anchors and produces causal local-motion tokens over multiple prediction horizons for online policy conditioning.
An action-group residual adapter incorporates these tokens only into the base branch.
Experiments use a 500-trajectory portable for each task, with 75 trajectories held out for RGB-VO evaluation, and 200 robot demonstrations as references.
Our policy, post-trained only on portable demonstrations, achieves 18.2\% lower base-velocity error than a policy trained with robot-collected demonstrations, while maintaining comparable end-effector translation accuracy.
Offline reconstruction reduces absolute trajectory errors for the body and hand streams by 24.6\% on average relative to the best evaluated baseline for each stream.
The complete system improves the mean success rate by 8.3 percentage points over OpenPI 0.5 across three real-robot tasks.
\end{abstract}

\section{Introduction}
Mobile manipulation is a foundational capability for embodied intelligence, enabling robots to interact with the physical world in applications ranging from domestic assistance and infrastructure inspection to industrial automation.
Mobile manipulation data collection typically involves long-horizon tasks, composed of multiple sequential stages, 
requiring a generalizable policy that can coordinate precise whole-body motion with fine-grained end-effector (EE) manipulation.
Recent whole-body vision-language-action models (VLAs) \cite{jiang2026wholebodyvla,hu2026openhlmempiricalrecipewholebody} and world action models (WAMs) \cite{li2026omega0latentpredictiveworld, zheng2026motionwamfoundationworldaction} advocate learning mobile manipulation as an integrated whole-body behavior rather than as separate navigation and arm-manipulation skills, but scaling such learning remains challenging due to the limited availability of demonstration data. 
Existing demonstration-collection paradigms often rely on teleoperation, which is expensive, slow, and difficult to scale for scene generalization. 
Cross-embodiment collection methods, such as UMI \cite{chi2024umi} and its variants \cite{FastUMI ,IPhUMI,xu2026hommi,huang2026mobileumi}, use handheld grippers with mounted cameras to collect manipulation demonstrations, 
while egocentric approaches use head-mounted cameras to capture broader scene context \cite{shi2026egohumanoid}. 
For mobile manipulation tasks, existing methods have shown that portable demonstration collection is feasible
\cite{ha2024umionlegs,huang2026mobileumi,xu2026hommi}. 
  
However, these approaches commonly depend on special devices, such as high-quality IMU sensors, GoPro cameras, iPhones, and VR headsets, 
which raise the hardware barrier for more accessible data collection and limit the number of demonstrators who can participate.
Using visual odometry (VO) can reduce the complexity of the data-collection hardware, but it remains vulnerable to drift over long trajectories and large rotations, which can degrade data quality or even make the data unusable.
Existing portable collection pipelines typically require accurate device calibration to recover reliable trajectories, placing strict demands on data post-processing, especially for mobile manipulation tasks.

\begin{figure}[t]
    \vspace{0.5em}
    \centering
    \includegraphics[width=\columnwidth]{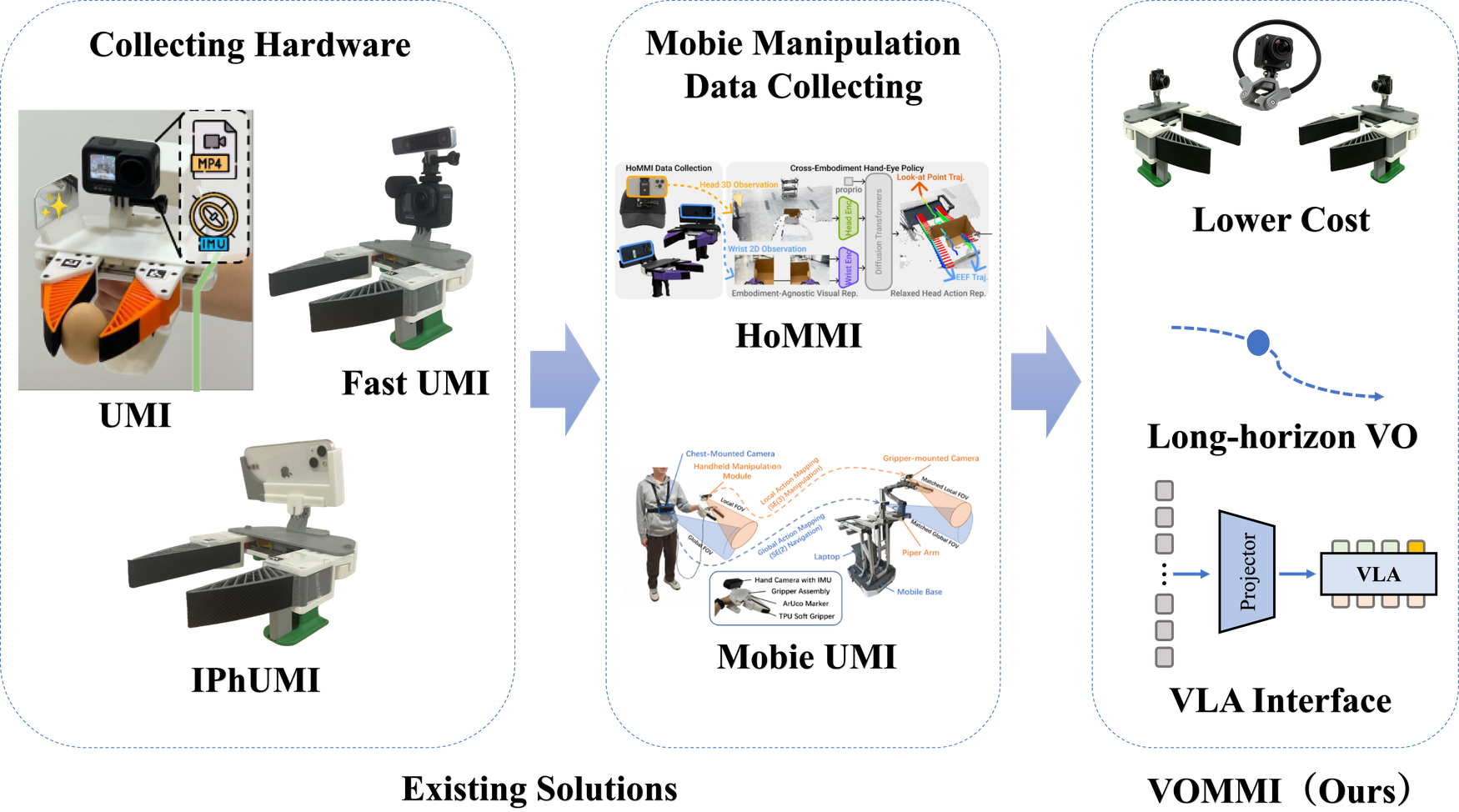}
    \caption{
    VOMMI.
    We build upon portable manipulation interfaces with a low-cost RGB-based perception system to enable accessible mobile demonstration collection and integration with VLA frameworks.
    }
    \label{fig:motivation}
    \vspace{-1.5em}
\end{figure}

As shown in Figure~\ref{fig:motivation}, we propose a VO-conditioned mobile-manipulation interface (VOMMI) to establish the data pipeline for
training whole-body VLA policies from portable demonstrations. 
VOMMI records RGB videos from body and hand views and uses VO to recover local motion cues, enabling low-cost demonstration collection.
Collection-time pose references are used only for offline metric supervision and evaluation, while deployment relies solely on RGB observations, camera intrinsics, timestamps, and robot proprioception.
To improve trajectory quality during offline data processing, VOMMI can leverage pretrained visual geometry for base movement.
These tokens are fed into the VLA post-training interface and connected only to the base-action residual adapter, while the manipulation
residual remains blind to body motion. This design provides the missing local visual proprioception for navigation-conditioned mobile manipulation without
claiming global localization or a new general-purpose VO algorithm. VOMMI is designed to test whether portable RGB demonstrations can improve long-horizon
mobile-manipulation competence in a VLA policy, while making the answer auditable by separating portable-data supervision from robot-data supervision. 
Our contributions are summarized as follows:
\begin{itemize}
    \item We propose a portable data-collection protocol for mobile manipulation with separate body and hand
    views, requiring no access to the target robot or human–robot kinematic correspondence calibration.
    \item  We present R2-VO, a real-time and refined RGB motion reconstruction module that produces low-latency
    local motion tokens for online VLA inference and anchor-refined trajectories for offline dataset processing.
    \item We define an action interface with dimension-specific reference pose
    supervision and action-group residuals, so that the manipulation residual remains
    independent of the body-motion token.
\end{itemize}

\section{Related Work}

\subsection{Portable Demonstration Interfaces}
Portable data collection is fast, cheap, and easy to deploy for mobile manipulators in diverse environments.
UMI \cite{chi2024umi} records in-the-wild manipulation with a portable
handheld interface and transfers relative EE actions to a robot. 
To enhance mobility, UMI-on-Legs combines this interface with a manipulation-centric whole-body controller \cite{ha2024umionlegs}, while
Mobile UMI \cite{huang2026mobileumi} introduces multiple views and decoupled planar-base and EE representations for mobile manipulation. HoMMI \cite{xu2026hommi} and EgoHumanoid \cite{shi2026egohumanoid} further use egocentric human demonstrations to capture whole-body task context. 
These systems demonstrate the value of extending portable whole-body data from fixed-base to mobile manipulation tasks, but they require customized, dedicated hardware.
In contrast, VOMMI addresses a complementary sensing question: how to recover the demonstrator's motion from RGB observations without requiring metric pose or IMU measurements as policy inputs.

\subsection{Robot Learning for Mobile Manipulation}
Mobile manipulation requires coordinated locomotion and manipulation over extended workspaces. 
HomeRobot integrates perception, navigation, and manipulation in realistic environments \cite{yenamandra2023homerobot}. 
Recent whole-body learning methods address complementary control challenges, from force-aware compensation for contact-rich loco-manipulation to latent policies for unified loco-manipulation \cite{jiang2026wholebodyvla,ding2025quarvlavisionlanguageactionmodelquadruped}.
In addition, bimanual manipulation learning uses egocentric demonstrations to learn active perception behaviors \cite{chuang2026lookfocusactefficient}.
DECOWAM~\cite{DECOWAM} trains adapters by separating base and arm latents and conditioning on base velocity to improve both future-video and action prediction.
Despite these advances, existing portable-demonstration-based mobile manipulation policies often lack an explicit representation of recent ego-motion when learning from portable RGB demonstrations. 
VOMMI addresses this gap by incorporating visual-odometry (VO) observations directly into the VLA and introducing separate residual pathways for body and manipulation actions. 
This design enables motion-aware whole-body correction without requiring metric pose or IMU measurements as policy inputs at deployment.

\subsection{Visual Body Motion for Policy State}

Learning-based VO systems such as DROID-SLAM and DPVO improve sequential motion estimation, while VGGT and VGGT-SLAM 2.0 extend feed-forward geometry
toward camera and real-time scene reconstruction \cite{teed2021droidslam,teed2022dpvo,wang2025vggt,maggio2026vggtslam2}.
Beyond geometric reconstruction, recent robot learning studies also explore
recovering actionable trajectories from human or egocentric demonstrations,
such as observation-only imitation and scalable egocentric data collection
\cite{karnan2021voilavisualobservationonlyimitationlearning,kareer2024egomimicscalingimitationlearning}. However, these approaches either
optimize geometric consistency or focus on embodiment alignment, rather than
extracting policy-oriented motion for mobile manipulation.

VOMMI considers visual motion from a policy-oriented perspective. 
This causal local representation reduces sensitivity to constant world-frame offsets while
preserving short-term ego-motion cues relevant for whole-body control. The
resulting token provides a compact visual proprioceptive state for policy
conditioning rather than a globally consistent trajectory or localization
memory.

\section{Data Collection System}
\subsection{Dual-view capture and temporal alignment}
\begin{figure}[h]
    \vspace {1.0em}
    \centering
    \includegraphics[width=\columnwidth]{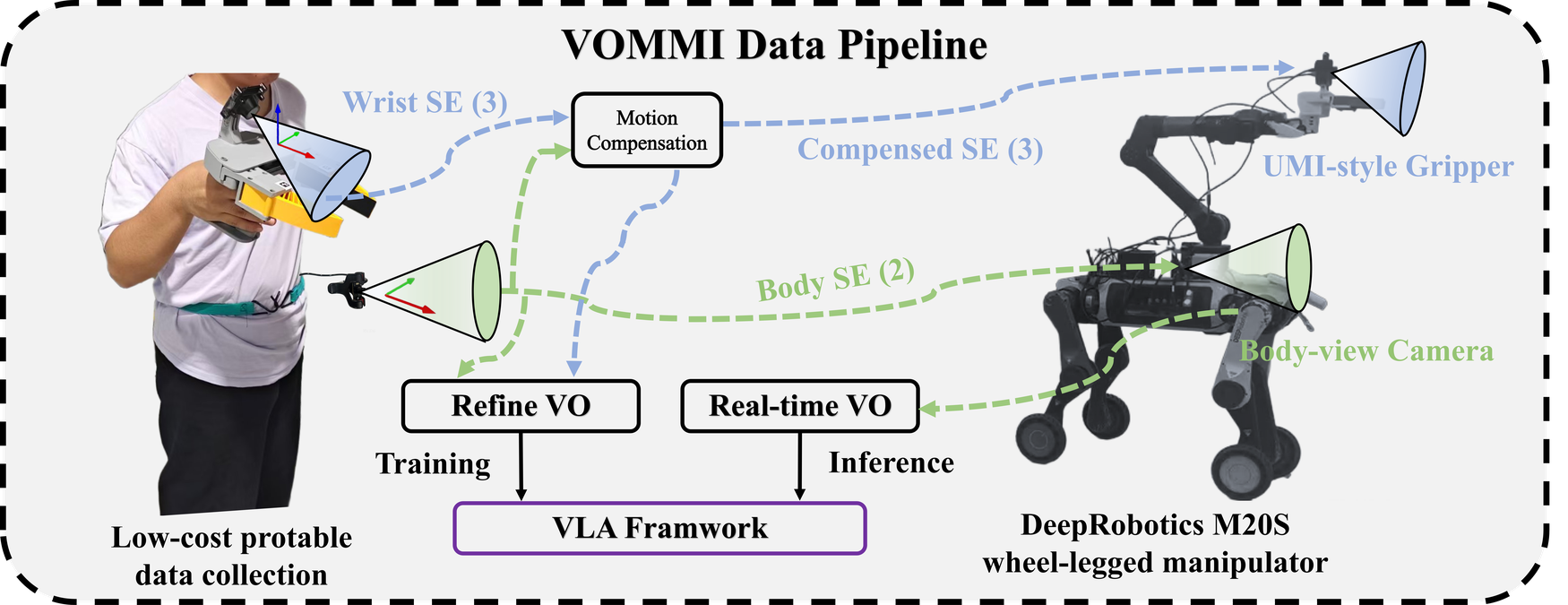}
    \caption{Portable data collection interface. Independent Body and Hand RGB
    views are timestamped into a robot-compatible episode.}
    \label{fig:pipeline}
\end{figure}

VOMMI is designed around a specific robot-free data-collection requirement: each demonstration must capture both the demonstrator’s locomotion and hand–object interaction using only RGB observations, without relying on additional sensing or target-robot instrumentation.
Fig.~\ref{fig:pipeline} summarizes the data flow.
The demonstrator wears a chest-mounted fisheye RGB camera for scene context and body motion and carries a second fisheye RGB camera on the UMI-style gripper module for interaction. 
These two views are recorded as complementary streams and reconciled later through local timestamps and trajectory-based compensation, resampled at 15~Hz after validity checks. 
Each episode stores the language instruction, both videos and intrinsics, tool center point (TCP) and gripper signals, and validity masks. 
The resulting episode contains both the observations required for RGB reconstruction and the action targets necessary to train a whole-body policy.
Collection-time sensor-derived trajectories serve as privileged offline references for R2-VO supervision and evaluation, but are not policy inputs at deployment.
As shown in Table~\ref{tab:hardware-cost}, the cost of our VOMMI's hardware is significantly lower than that of current portable data acquisition methods.

\begin{table}[t]
\centering
\vspace{0.5em}
\caption{Representative hardware requirements and estimated acquisition cost
for portable data collection methods}
\label{tab:hardware-cost}
\scriptsize
\begin{tabularx}{\columnwidth}{@{}l l l c X@{}}
\toprule
Method & Camera(s) & View & Price (USD) \\
\midrule
UMI~\cite{chi2024umi} & GoPro & Wrist & \$350--450  \\
iPhUMI~\cite{IPhUMI} & iPhone & Wrist & \$1{,}000--1{,}200  \\
FastUMI~\cite{FastUMI} & GoPro+RealSense T265& Wrist & \$650--900  \\
Mobile UMI~\cite{huang2026mobileumi} & Dual RGB & Wrist+ego & \$800--1{,}200  \\
VOMMI (ours) & 2 fisheye RGB & Wrist+ego & \$100--200 \\
\bottomrule
\end{tabularx}
\vspace{-2em}
\end{table}

\subsection{Episode schema and split protocol}
The collection contains 500 portable episodes per evaluated task. For each task, 75 complete episodes form a fixed held-out RGB-VO evaluation split and are excluded from estimator fitting, canonicalizer fitting, and model selection.
The remaining 425 episodes support model development and downstream VLA training under the experimental protocol.

For each valid frame in the Hand stream, let
$T^{W_H}_{E}(t)\in\SEthree$ denote the TCP pose at time $t$,
expressed in the Hand-stream local world frame $W_H$.
The current-relative manipulation target at future step $k$
is then defined as:
\begin{equation}
A^{\mathrm{tcp}}_{t,k}
=\left(T^{W_H}_{E}(t)\right)^{-1}T^{W_H}_{E}(t+k+1),
\label{eq:tcp-relative-label}
\end{equation}
which is the tracking-derived current-TCP-relative $\SEthree$ reference used
for Hand supervision. The Body stream supplies the corresponding planar reference
$[\Delta x,\Delta y,\Delta\psi]$. During RGB source replacement, the canonicalized Body trajectory is aligned
with the Hand timestamps and used to compensate the Hand TCP motion. Section~V
defines this compensation and the resulting action targets.

\section{Problem Formulation}

The central data-construction problem is to recover a robot-compatible local motion representation from two unsynchronized RGB streams, enabling the transfer of portable human demonstrations to robot policy learning.
Let $I^B_{0:T}$ and $I^H_{0:T}$ denote the Body and Hand videos. 
The Body trajectory is represented by planar motion as:
\begin{equation}
\xi^B_{t,k}=[\Delta x_{t,k},\Delta y_{t,k},\Delta\psi_{t,k}]\in\mathfrak{se}(2),
\label{eq:body-reference}
\end{equation}
whereas the Hand trajectory captures fine-grained manipulation motion through current-relative TCP motion, yielding a local SE(3) representation that reduces dependence on the global reference frame:
\begin{equation}
\xi^E_{t,k}=\log\!\left(\mathcal T_E(t)^{-1}\mathcal T_E(t+k+1)\right)
=[\Delta p_{t,k},\phi_{t,k}]\in\mathfrak{se}(3).
\label{eq:hand-reference}
\end{equation}

We estimate the motions in Eqs.~\eqref{eq:body-reference}
and~\eqref{eq:hand-reference} from past RGB observations and compensate the
Hand motion using the planar Body estimate. The Body trajectory is
canonicalized and aligned with the Hand timestamps, yielding
Eq.~\eqref{eq:problem-compensation}:
\begin{equation}
T^{\mathrm{comp}}_{E|B}(t)
=\left(T_B(t)\right)^{-1}T_E(t),
\label{eq:problem-compensation}
\end{equation}
where $T_B(t)$ and $T_E(t)$ are expressed in the same reference frame.
Reference trajectories provide privileged supervision for offline data construction
and evaluation, while deployment relies only on RGB observations.

\subsection{Unified Observation and Action Space}

The source-conditioned observation is:
\begin{equation}
\mathcal O_t^s=(\ell,I_{t-h}^{B,s},I_{t-h}^{H,s},s_t^s,M_t^o),
\label{eq:observation}
\end{equation}
In Eq.~\eqref{eq:observation}, $M_t^o$ masks unavailable source-specific observation components.
Here, $t$ denotes the current policy frame, $k$ denotes a future step within the action chunk, and $H$ denotes the chunk horizon.
Superscripts $B$, $H$, and $E$ denote Body-camera, Hand-camera, and TCP quantities, respectively.
Superscripts ref'' and rgb'' denote collection-time reference and RGB-reconstructed trajectories, respectively.
The policy predicts a horizon-$H$ action chunk:
\begin{equation}
A_{t+H-1}\in\R^{H\times16}, \qquad H=50,
\label{eq:action-chunk}
\end{equation}
with each step represented canonically as:
\begin{equation}
 \begin{aligned}
 a_{t,k}&=[v_x,v_y,\dot\psi,\ \xi^E_{t,k},\ q_{0:5},g],\\[-2pt]
 \xi^E_{t,k}&=\log\!\left(T^{E_t}_{E_{t+k+1}}\right)
 =[\Delta p_{t,k},\phi_{t,k}]\in\R^6.
\end{aligned}
\label{eq:action}
\end{equation}
where $k=0,\ldots,H-1$, $\phi_{t,k}$ is the rotation-vector logarithm,
$q_{0:5}$ are the six arm-joint targets, and $g$ is the gripper target.
Eq.~\eqref{eq:action-chunk} sets a 50-step chunk at the 15-Hz policy clock,
and Eq.~\eqref{eq:action} expresses base velocities in the current Body frame
and $\xi^E_{t,k}$ in the current TCP frame.
The action mask $M_t^a \in {0,1}^{H \times 16}$ indicates the valid action dimensions.
The same masks are used in Sec.~V and exclude missing values rather than treating them
as zeros.

\section{VOMMI}

\begin{figure*}[htb]
    \centering
    \vspace{0.2em}
    \includegraphics[width=\textwidth]{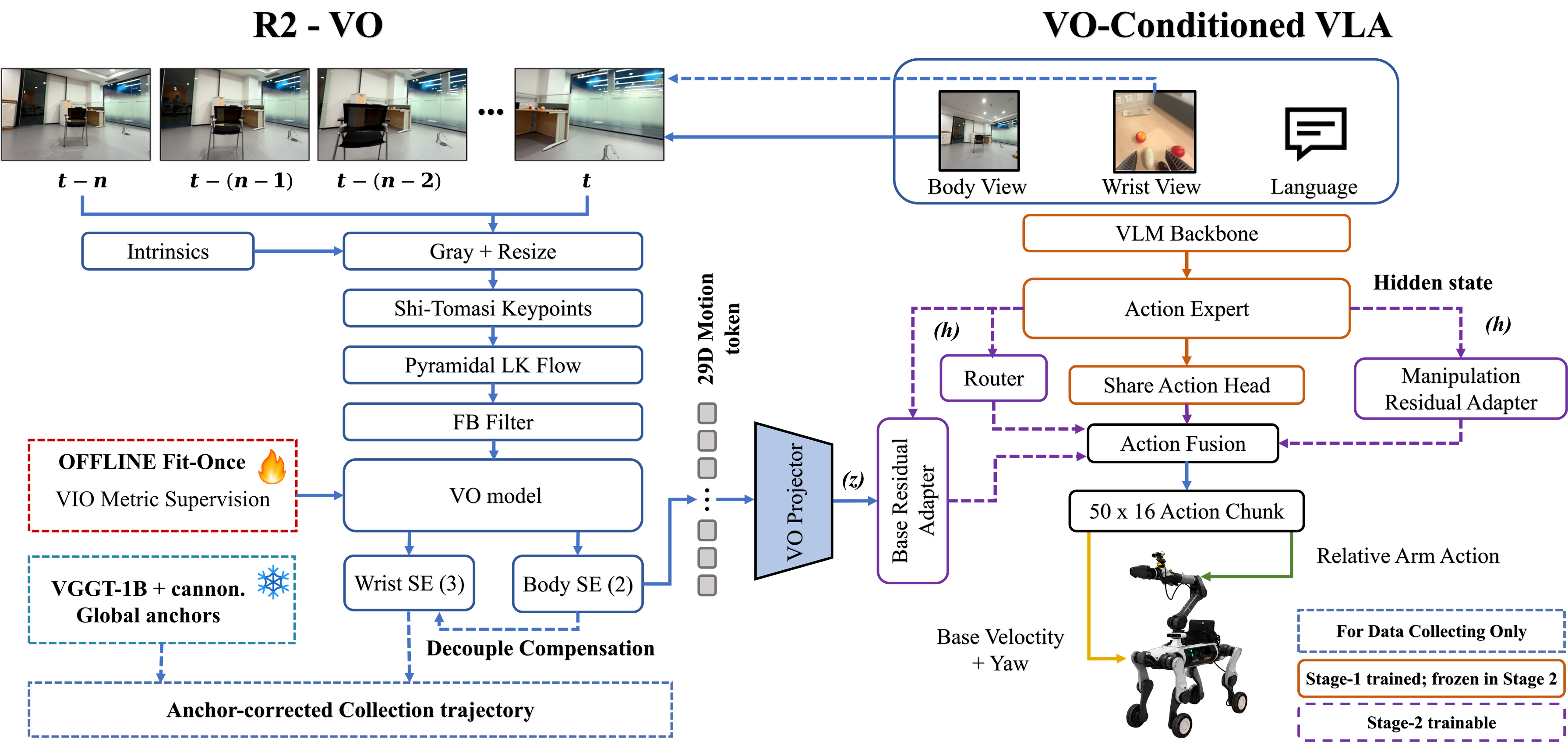}
    \caption{Overview of VOMMI. The online branch estimates multi-horizon local
    motion from RGB and conditions the base-action residual. The offline branch
    combines the causal estimator with frozen VGGT anchors, initializes
    the Body trajectory at its first valid pose, and applies planar Body
    compensation to the Hand stream before forming demonstration targets.
    Reference pose supervision available only during data collection.}
    \label{fig:framework}
    \vspace{-2em}
\end{figure*}
VOMMI converts portable RGB demonstrations into a unified whole-body policy through the pipeline shown in Fig.~\ref{fig:framework}. Local base and EE actions provide a common representation across demonstrator and robot embodiments. R2-VO refines demonstration trajectories offline using sparse geometric anchors and provides causal body-motion estimates from RGB observations online. We encode these online estimates as a multi-horizon motion token and integrate it into the policy through an action-group residual adapter, restricting motion-conditioned updates to the body actions.

\subsection{Local Cross-Embodiment Action Representation}
We express body motion commands in the current body frame and EE targets in the current TCP frame.
This representation provides a common action interface across portable and robot data while avoiding supervision tied to an arbitrary world origin.
A per-sample mask, $M_i^a \in \{0,1\}^{H \times 16}$, selects the action dimensions available from each source, and the flow loss is normalized over the valid dimensions before being averaged across the batch.

Local actions reduce sensitivity to global pose errors. Let $T_t$ be the true
pose and $\widehat{T}_t = G_t T_t$ its estimate, where $G_t$ is a
time-varying world-frame error.
We define the current-frame relative target as:
\begin{equation}
\widehat{\Delta T}_{t,i}=
\widehat T_t^{-1}\widehat T_{t+i}
=T_t^{-1}G_t^{-1}G_{t+i}T_{t+i},
\label{eq:relative_error}
\end{equation}
Eq.~\eqref{eq:relative_error} cancels constant global translation and yaw
offsets when $G_t=G_{t+i}$. Otherwise, only the local error increment
$G_t^{-1}G_{t+i}$ enters the relative target.
The same property applies independently to the Body and Hand streams, even though their tracking frames are not mutually registered.
Accordingly, we formulate relative targets independently within each stream.

\subsection{Body-Hand Alignment and Planar Compensation}
\label{sec:base-motion-compensation}
A body-relative hand trajectory requires the hand TCP and body pose to
be expressed in a common coordinate system. 

Let $W_H$ and $W_B$ denote the tracking worlds of the hand and Body streams, $C_H$ and $C_B$ their camera
frames, $E$ the hand TCP, and $B$ the demonstrator body frame. 
For a setting with a verified rigid cross-device registration $A_{BH}=T^{W_B}_{W_H}$ and fixed camera extrinsics
$X_H=T^{C_H}_{E}$, $X_B=T^{C_B}_{B}$, the trajectories would satisfy:
\begin{align}
T^{W_H}_{E}(t) &= T^{W_H}_{C_H}(t)X_H,\label{eq:hand-extrinsic} \\
T^{W_B}_{B}(t) &= T^{W_B}_{C_B}(t)X_B,\\
\hat T^{W_B}_{E}(t) &= \hat A_{BH}\,T^{W_H}_{E}(t).
\label{eq:cross-device-registration}
\end{align}

Applying Eq.~\eqref{eq:problem-compensation} to the registered poses in
Eqs.~\eqref{eq:hand-extrinsic}--\eqref{eq:cross-device-registration} gives
the body-relative TCP pose:
\begin{equation}
T^{B}_{E}(t)=\bigl(T^{W_B}_{B}(t)\bigr)^{-1}
\hat T^{W_B}_{E}(t).
\label{eq:body-relative-pose}
\end{equation}

For RGB-based Body trajectory reconstruction, we refine the planar motion
estimated from RGB observations using a frozen VGGT+anchor branch.
Let $\widetilde{T}^{R}_{B}(t)\in\SEtwo$ denote the resulting canonicalized
Body trajectory, where VGGT provides global geometric constraints and sparse
anchors recover metric consistency to mitigate long-term VO drift.
We initialize this trajectory at its first valid timestamp $t_0$ as:
\begin{equation}
G_{\mathrm{init}}
=\widehat T^{R}_{B}(t_0)
\left(\widetilde T^{R}_{B}(t_0)\right)^{-1},
\qquad
\widehat T^{R}_{B}(t)
=G_{\mathrm{init}}\widetilde T^{R}_{B}(t).
\label{eq:rgb-initialized-body}
\end{equation}
The trajectory initialized in Eq.~\eqref{eq:rgb-initialized-body} supplies
planar displacement for compensation. Under a constant left gauge, its relative
increments are invariant as shown in Eq.~\eqref{eq:body-gauge-cancellation}:
\begin{equation}
\widetilde T_B(t)^{-1}\widetilde T_B(t+k)
=
T_B(t)^{-1}T_B(t+k).
\label{eq:body-gauge-cancellation}
\end{equation}

Planar compensation projects the Body motion to $\SEtwo$ and lifts it to $\SEthree$:
\begin{equation}
T^{W_B}_{B,\parallel}(t)=
\begin{bmatrix}
R_z(\psi_B(t)) & [x_B(t),y_B(t),0]^\top\\
\mathbf 0^\top & 1
\end{bmatrix},
\label{eq:planar-body-pose}
\end{equation}
where $(x_B,y_B,\psi_B)$ are the body translation and yaw in the common
reference. Using Eq.~\eqref{eq:planar-body-pose} in
Eq.~\eqref{eq:body-relative-pose} gives the planar-compensated Hand pose:
\begin{equation}
\bar T_{E|B}^{\parallel}(t)=
\bigl(T^{W_B}_{B,\parallel}(t)\bigr)^{-1}
\hat T^{W_B}_{E}(t),
\label{eq:planar-compensated-pose}
\end{equation}
and the current-relative future target is:
\begin{equation}
A^{ee,\parallel}_{t,k}=
\left(\bar T_{E|B}^{\parallel}(t)\right)^{-1}
\bar T_{E|B}^{\parallel}(t+k+1).
\label{eq:compensated-ee-target}
\end{equation}

For $q\in\{\mathrm{ref},\mathrm{rgb}\}$, let $\mathbf p_B^q(t)$ and $\psi_B^q(t)$ denote the canonical Body translation and heading. 
Here, $q=\mathrm{ref}$ denotes the collection-time reference and $q=\mathrm{rgb}$ denotes the RGB reconstruction.
After interpolation to hand timestamps, we form the planar body transform:
\begin{equation}
T^{W_B}_{B,q}(t)=
\begin{bmatrix}
R_z(\psi_B^q(t)) & [p^q_{B,x}(t),p^q_{B,y}(t),0]^\top\\
\mathbf 0^\top & 1
\end{bmatrix}.
\label{eq:source-dependent-compensation}
\end{equation}
The body-normalized TCP pose and relative target are:
\begin{equation}
\begin{aligned}
\bar T_{E|B}^{q}(t)&=
  \left(T^{W_B}_{B,q}(t)\right)^{-1}\hat T^{W_B}_{E}(t),\\
A^{ee,q}_{t,k}&=
  \left(\bar T_{E|B}^{q}(t)\right)^{-1}
  \bar T_{E|B}^{q}(t+k+1).
\end{aligned}
\label{eq:source-dependent-ee-target}
\end{equation}
Portable RGB source-replacement and VLA experiments use
Eqs.~\eqref{eq:source-dependent-compensation}--\eqref{eq:source-dependent-ee-target}
with $q=\mathrm{rgb}$ to compensate Body translation in $x$ and $y$ and Body yaw.
These equations apply the planar-compensation rule in
Eqs.~\eqref{eq:planar-compensated-pose}--\eqref{eq:compensated-ee-target}
to the selected Body trajectory source.

During deployment, the robot-side adapter composes the predicted base transform
$T^{W_R}_{B,\mathrm{cmd}}(t+k)$ and emits:
\begin{equation}
T^{E_t}_{E_{t+k},\mathrm{cmd}} =
  \bigl(T^{W_R}_{E}(t)\bigr)^{-1}
T^{W_R}_{B,\mathrm{cmd}}(t+k)\,
\bar T_{E|B}^{\mathrm{des}}(t+k),
\label{eq:body-to-tcp-command}
\end{equation}
where $\bar T_{E|B}^{\mathrm{des}}$ denotes the planar-compensated target.
Eq.~\eqref{eq:body-to-tcp-command} yields the six current-TCP-relative EE
dimensions.

\subsection{R2-VO}

R2-VO provides local metric motion from RGB in two complementary modes.
The causal branch produces motion tokens for deployment, while the offline
branch refines trajectories during demonstration construction. Both branches
share the same frozen motion estimator.

\subsubsection{Causal Local Motion Estimation}

For consecutive Body-camera frames, the collection-time tracker provides the
planar motion target
$\Delta\xi_{t,1}=[\Delta x,\Delta y,\Delta\psi]$.
We extract visual motion descriptors from optical flow, geometric cues, and
track quality statistics, resulting in a 206-dimensional descriptor.
A seven-frame temporal window augments this descriptor to 317 dimensions by
incorporating temporal motion information.
An extremely randomized tree regressor with 32 trees predicts the one-step
motion target. During inference, the estimator uses only past RGB frames,
camera intrinsics, and timestamps.

Atomic predictions are resampled to the 15-Hz clock and composed over horizons of $k\in\{2,3,6,12\}$ frames, corresponding to ${0.13,0.20,0.40,0.80}$~s.
Starting from $(x,y,\theta)=(0,0,0)$, each step integrates the body-frame velocities $[v_x,v_y,\omega]$ as follows:
\begin{align}
\theta_{j+\frac12}&=\theta_j+\tfrac12\omega_j\Delta t,\label{eq:integrate-midpoint}\\
x_{j+1}&=x_j+(v_{x,j}\cos\theta_{j+\frac12} -v_{y,j}\sin\theta_{j+\frac12})\Delta t,\\
y_{j+1}&=y_j+(v_{x,j}\sin\theta_{j+\frac12} +v_{y,j}\cos\theta_{j+\frac12})\Delta t,\\
\theta_{j+1}&=\theta_j+\omega_j\Delta t,
\label{eq:integrate}
\end{align}
where $\Delta t=1/15$. Each horizon is represented by:
\begin{equation}
 [x,y,\sin\theta,\cos\theta,
   \log\sigma_x,\log\sigma_y,\log\sigma_\theta],
 \label{eq:horizon-token}
\end{equation}
with $\sigma_x=\sigma_y=\sigma_\theta=0.02\sqrt{k/2}$. 
Eqs.~\eqref{eq:integrate-midpoint}--\eqref{eq:integrate} compose
source-Body-frame increments in each horizon's window-start frame.
Concatenating four encodings from Eq.~\eqref{eq:horizon-token} with a
history-validity bit yields $\nu_t\in\R^{29}$. The incomplete history is assigned $\nu_t=0$ and $q_t=0$. 
The resulting token encodes causal short-horizon Body motion.

\subsubsection{Offline Drift Refinement}

For offline action construction, R2-VO combines the integrated causal
trajectory with 1-Hz VGGT-1B anchors~\cite{wang2025vggt}. A low-capacity
canonicalizer $C$ maps VGGT coordinates to the metric initial-camera frame.
Let $g_{t,xy}$ denote the sparse metric anchors obtained from
canonicalized VGGT geometry, $c_{t,xy}$ denote the integrated causal
VO trajectory, and $L(c_{xy})_t$ denote the causal trajectory evaluated
at the anchor timestamps. The refined trajectory preserves short-term
VO motion while correcting accumulated drift through sparse geometric constraints:
\begin{equation}
    \hat{p}^{\mathrm{off}}_{t,xy}
    =
    g_{t,xy}
    +
    c_{t,xy}
    -
    L(c_{xy})_t .
    \label{eq:offline-anchor}
    \end{equation}

The canonicalizer is a weighted Ridge map from first-frame-centered VGGT
positions and temporal features to metric reference positions in the
initial-camera frame:
\begin{equation}
\min_{\Theta}\sum_e\frac{1}{N_e}\sum_t
\left\|
\begin{bmatrix}
(\phi(u_{e,t})-\mu)/s\\1
\end{bmatrix}\Theta-p^{\mathrm{ref}}_{e,t}
\right\|_2^2+\lambda\|\Theta\|_F^2,
\label{eq:canonicalizer}
\end{equation}
In Eq.~\eqref{eq:canonicalizer}, $u_{e,t}$ is the normalized VGGT position,
and $\phi$ supplies linear, time-augmented, or quadratic features for metric
translation prediction.
For the Hand stream, sparse anchors refine the vertical component during offline action
construction, while the Hand VO branch predicts the complete local TCP
$\SEthree$ increment, including rotation, from RGB observations. 
The TCP-relative pose in Eq.~\eqref{eq:tcp-relative-label} serves as the training and evaluation target.

For source-replaced demonstrations, the offline Body trajectory from
Eq.~\eqref{eq:offline-anchor} is interpolated to the Hand timestamps and
inserted into Eq.~\eqref{eq:source-dependent-compensation} to construct
demonstration labels. Online inference uses causal local increments.

\subsection{VO-Conditioned Action-Group Adaptation}

To condition the base actions on visual motion, we augment the pretrained
action expert with separate residual branches for base and manipulation actions.
Let $h_{t,k}\in\R^{1024}$ and $u^0_{t,k}\in\R^{32}$ denote the expert
hidden state and flow prediction at step $(t,k)$. 
A LayerNorm-MLP $P$ projects the motion token, centered and gated by a validity indicator:
\begin{equation}
z_t=q_t[P(\nu_t)-P(0)].
 \label{eq:center}
\end{equation}
After lifting $z_t$ from Eq.~\eqref{eq:center} to the hidden dimension, the base residual is:
\begin{equation}
 r^b_{t,k}=R_b([h_{t,k}\odot W_z z_t,z_t])-R_b(0),
 \label{eq:base-residual}
\end{equation}
while the manipulation residual $r^m_{t,k}=R_m(h_{t,k})$ does not receive body motion. Independent gates $(g^b,g^m)=\sigma(W_rh_{t,k})$ combine the two paths:
\begin{equation}
 \widehat u_{t,k}=u^0_{t,k}
 +g^b_{t,k}r^b_{t,k}M_b
 +g^m_{t,k}r^m_{t,k}M_m,
 \label{eq:residual}
\end{equation}
where $M_b$ selects dimensions $0{:}3$, $M_m$ selects $3{:}16$, and both are zero on padded dimensions. 
Zero-initialized outputs preserve the pretrained policy at initialization.
Eqs.~\eqref{eq:base-residual}--\eqref{eq:residual} restrict
motion-conditioned updates to the base group.
Router targets are derived from base and manipulation activity.
The training objective is:
\begin{equation}
\mathcal L=\frac{1}{B}\sum_i\mathcal L_{\mathrm{flow}}^{(i)}
 +0.1\,\mathcal L_{\mathrm{router}},
 \label{eq:loss}
\end{equation}
where Eq.~\eqref{eq:loss} averages the validity-normalized per-sample flow
losses. Each flow loss applies the corresponding action masks.

\section{Experiments}
In this section, we evaluate the impact of R2-VO post-processing on data quality, the performance of the VO-conditioned VLA policy, and real-world deployment across diverse mobile manipulation tasks.
The experiments address the following research questions:
\begin{itemize}
\item \textbf{Q1:} How accurately can R2-VO reconstruct tracking-derived
planar $\SEtwo$ and local TCP $\SEthree$ pose references from RGB data?
\item \textbf{Q2:} Do VO-refined data improve real-world rollouts?
\item \textbf{Q3:} How does the local motion token affect long-horizon mobile manipulation?
\item \textbf{Q4:} Which components contribute most significantly to the R2-VO and VOMMI?
\end{itemize}

\subsection{Experimental Setting}
\textbf{Tasks and evaluation metrics.}
Our benchmark contains three mobile-manipulation tasks: moving across barriers
to pick up different fruits, moving in and out of a room to manipulate a
bottle, and opening a cabinet door to place a folded box. We use
collection-time tracking-derived references for both streams: a planar metric
motion reference for Body and a tracking-derived local TCP $\SEthree$ pose
reference for Hand. For data post-processing, we report trajectory error against these
references. 
For real-world rollout, we evaluate the wheel-legged DeepRobotics M20S equipped with a CM1 manipulator over 20 independent trials per condition, reporting task success rate and action deviation from the reference commands.

\textbf{Baselines.}
To evaluate VOMMI, we compare it with DROID-W~\cite{teed2021droidslam}, DPVO~\cite{teed2022dpvo}, MASt3R-SLAM~\cite{murai2025mast3rslamrealtimedenseslam}, and
CUT3R~\cite{wang2025continuous3dperceptionmodel}, and VGGT-SLAM 2.0~\cite{maggio2026vggtslam2}
for VO performance, as well as OpenPI 0.5~\cite{black2025pi05}, GR00T N1.6~\cite{grootn1}, and X-VLA~\cite{XVLA} for real-world evaluation.

\subsection{Main Experiment Results}
\textbf{VO motion accuracy (Q1).}
We evaluate the frozen RGB-based R2-VO on the fixed 75-episode held-out test
split defined above. Table~\ref{tab:rgbvo} reports the
per-step relative-motion reconstruction error for the Body and Hand streams in their local references.
Source-replacement and VLA use the RGB-source planar compensation ($q=\mathrm{rgb}$) defined in Sec.~\ref{sec:base-motion-compensation}. Body-span uses the same held-out split.

\begin{table*}[t]
\centering
\vspace{1em}
\caption{Local and accumulated motion errors on VOMMI trajectories. Lower is better.}
\label{tab:rgbvo}
\label{tab:vla-relative-action}
\scriptsize
\setlength{\tabcolsep}{1.5pt}
\begin{tabular}{@{}lcccc@{\hspace{6pt}}cccccc@{}}
\toprule
& \multicolumn{4}{c}{Body $\SEtwo$}
& \multicolumn{6}{c}{Hand $\SEthree$} \\
\cmidrule(lr){2-5}\cmidrule(lr){6-11}
& \multicolumn{2}{c}{Local (mean/P90)}
& \multicolumn{2}{c}{Accumulated}
& \multicolumn{2}{c}{Local (mean/P90)}
& \multicolumn{4}{c}{Accumulated} \\
\cmidrule(lr){2-3}\cmidrule(lr){4-5}\cmidrule(lr){6-7}\cmidrule(lr){8-11}
Method & Trans. (cm) & Yaw ($^\circ$) & XY ATE (cm) & XY End (cm)
& Trans. (cm) & Rot. ($^\circ$) & 3D ATE (cm) & 3D End (cm) & XY ATE (cm) & XY End (cm) \\
\midrule
Ours (online) & 0.26/0.54 & \textbf{0.092}/0.227 & 15.65 & 19.29
& 0.60/\textbf{1.08} & 0.346/0.717 & 71.82 & 60.42 & 71.51 & 60.30 \\
Ours (offline) & \textbf{0.24/0.58} & 0.093/\textbf{0.224} & \textbf{5.97} & \textbf{5.33}
& \textbf{0.60}/1.08 & \textbf{0.340/0.704} & \textbf{50.47} & \textbf{56.68} & \textbf{50.23} & \textbf{56.66} \\
DPVO & 1.24/2.86 & 0.412 0.537& 28.09 & 27.86
& 14.48/21.90$^\dagger$ & 3.887/8.591$^\dagger$ & 220.28 & 219.21 & 220.16 & 219.21 \\
DROID-SLAM & 1.50/3.21 & 0.309/0.662 & 83.86 & 62.67
& 2.85/4.62$^\dagger$ & 2.358/4.834$^\dagger$ & 124.54 & 109.75 & 124.16 & 109.45 \\
DROID-W & 0.45/1.07 & 0.140/0.345 & 8.54 & 8.02
& 4.91/7.83$^\dagger$ & 3.202/6.665$^\dagger$ & 126.12 & 138.17 & 125.80 & 137.86 \\
VGGT-SLAM 2.0 & 0.86/1.63 & 0.227/0.463 & 6.88 & 9.95
& 10.95/19.44$^\dagger$ & 6.100/10.443$^\dagger$ & 95.41 & 83.90 & 94.78 & 83.68 \\
CUT3R & 24.52/52.81 & 0.906/1.954 & 23.91 & 25.01
& 23.69/41.60$^\dagger$ & 10.748/23.975$^\dagger$ & 83.35 & 103.57 & 83.02 & 103.57 \\
MASt3R-SLAM & 63.71/106.80 & 0.581/1.137 & 66.03 & 57.43
& 165.25/235.28$^\dagger$ & 35.939/59.034$^\dagger$ & 78.88 & 84.80 & 78.42 & 85.31 \\
\bottomrule
\end{tabular}
\end{table*}

At the deployed 15-Hz step, the causal Ours branch gives 0.26/0.54~cm
Body translation error (mean/P90). Its TCP-relative Hand-stream evaluator gives
0.601/1.081~cm translation and 0.346/0.717$^\circ$ $\SOthree$ rotation. With
complete-sequence access for demonstration conversion, the Ours Offline
(VGGT+anchor) branch gives 0.244/0.592~cm and 0.093/0.224$^\circ$ for Body,
and 0.600/1.082~cm and 0.340/0.704$^\circ$ for hand at the same 67-ms
reference span. On an Intel Core i5-9600KF CPU, the causal Body branch takes
10.02~ms on average (P95: 13.76~ms), excluding image capture and decoding.
Table~\ref{tab:rgbvo} uses local increments:
Body is planar $\SEtwo$ and Hand is TCP-relative $\SEthree$. Ours and DPVO use
matched stored-frame spans, whereas CUT3R and MASt3R-SLAM retain native sparse
steps from 1-Hz RGB bundles. Thus, the baseline rows are not all 15-Hz runs.

At the deployed 15-Hz step, R2-VO lowers mean translation error by 80.4\% and
83.8\% relative to DPVO and DROID-SLAM, respectively. At 267~ms, it retains
0.878~cm mean error, corresponding to reductions of 79.9\% and 84.1\%. The
measured latency is compatible with online 15-Hz inference. Accumulated
trajectory drift is reported in the accumulated-error columns of Table~\ref{tab:rgbvo}.

Table~\ref{tab:rgbvo} groups local and accumulated errors by stream. Bold entries
indicate minima before rounding. Integrating the causal Body branch gives 15.65~cm ATE
and 19.29~cm endpoint error. The offline VGGT+anchor variant reduces these to
5.97~cm and 5.33~cm. The accumulated metrics are therefore trajectory
diagnostics rather than deployment-rate runtime rankings. Offline reconstruction
reduces mean absolute Body/Hand trajectory error by 24.6\% versus the best
baseline for each stream.

\begin{figure}[t]
    \centering
    \vspace {1.0em}
    \includegraphics[width=\columnwidth]{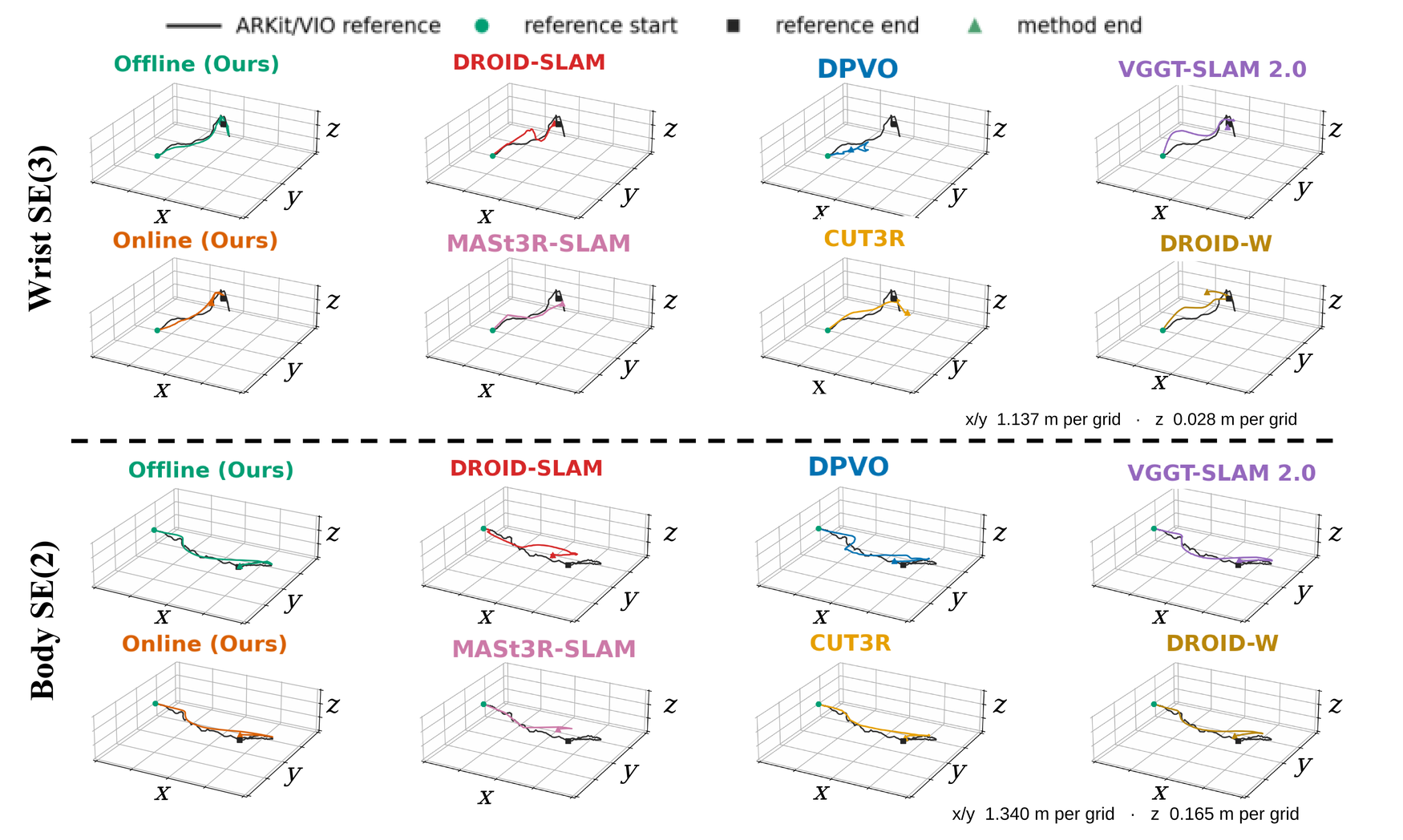}
    \caption{
        Camera trajectories reconstruction on held-out episode trail. The online stream and reference are projected to the ground plane. Offline Body scores in Table~\ref{tab:vla-relative-action} use the common $x$--$y$ state.
    }
    \label{fig:vo-baseline-episode57}
    \vspace {-1.0em}
\end{figure}

\begin{figure}[!htb]
    \centering
    \includegraphics[width=\columnwidth]{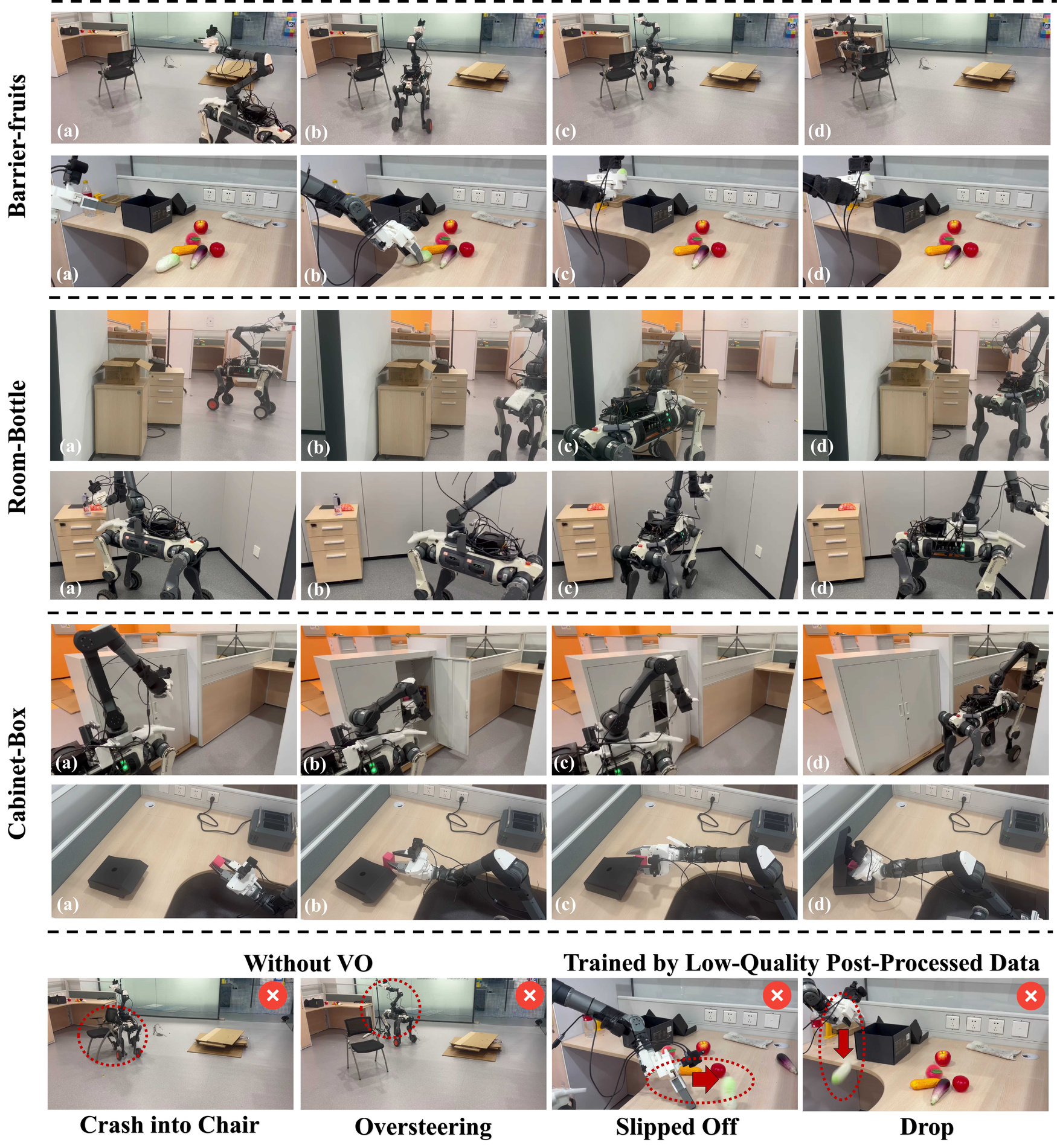}
    \vspace {-0.5em}    
    \caption{Qualitative M20S rollouts on three long-horizon mobile-manipulation
    tasks. The upper rows show representative stages from navigation and
    approach to object interaction and task completion. The bottom row shows representative failure cases, marked with red crosses.}
    \label{fig:robot-deployment}
\end{figure}

\begin{table}[!ht]
\caption{Complete-task success rate (\%) over 20 trials per task.}
\label{tab:task-success}
\centering
\scriptsize
\setlength{\tabcolsep}{2.5pt}
\begin{tabular}{@{}lccc@{}}
\toprule
Method & Barriers--fruit & Room--bottle & Cabinet--box \\
\midrule
OpenPI 0.5 & 35 & 50 & \textbf{65} \\
GR00T N1.6 & 20 & 25 & 10 \\
X-VLA & 25 & 35 & 30 \\
VOMMI (ours) & \textbf{65} &  \textbf{60} & 50 \\
\bottomrule
\end{tabular}
\vspace {-1.0em}
\end{table}

\textbf{Overall performance on mobile manipulation tasks (Q2).}
Figure~\ref{fig:robot-deployment} shows representative rollouts.
Across the tasks in Table~\ref{tab:task-success}, mean success is 58.3\% for
VOMMI versus 50.0\% for OpenPI 0.5, an 8.3-percentage-point gain.
Table~\ref{tab:real-action-agreement} evaluates policy predictions against
reference actions on 15 real-robot trajectories. M20-trained uses robot
demonstrations. 
VOMMI-R/A use portable demonstrations with compensated current-TCP-relative targets and uncompensated absolute targets.
Metrics pool valid actions over 50-step horizons: Base MAE averages
$|v_x-v_x^*|$ and $|v_y-v_y^*|$. Yaw MAE averages
$|\dot\psi-\dot\psi^*|$, where stars denote reference commands.
EE errors compare forward-kinematic TCP poses in the current TCP frame.
VOMMI-R reduces base MAE from 0.044 to 0.036~m/s (18.2\%) relative to
M20-trained. Compared with VOMMI-A, EE translation/rotation errors fall from
0.569~cm/$1.091^\circ$ to 0.170~cm/$0.453^\circ$.

\begin{table}[h]
\vspace {-2em}
\caption{Open-loop action prediction errors.}
\label{tab:real-action-agreement}
\centering
\scriptsize
\setlength{\tabcolsep}{2.5pt}
\begin{tabularx}{\columnwidth}{@{}Xccccc@{}}
\toprule
Policy & \shortstack{Base MAE\\(m/s)}
& \shortstack{Yaw MAE\\(rad/s)}
& \shortstack{EE trans.\\mean (cm)}
& \shortstack{EE rot.\\mean ($^\circ$)}
& \shortstack{EE trans.\\max (cm)}
 \\
\midrule
M20-trained & 0.044 & 0.107 & \textbf{0.155} & \textbf{0.372} & \textbf{4.32}  \\
VOMMI-R & \textbf{0.036} & \textbf{0.104} & 0.170 & 0.453 &  4.55 \\
VOMMI-A & 0.079 & 0.204 & 0.569 & 1.091 & 13.96 \\
\bottomrule
\end{tabularx}
\vspace {-2em}
\end{table}


\subsection{Ablation Study}

\begin{table}[h]
\vspace {0.5em}
\caption{Residual ablation: flow loss ($\times10^{-2}$).}
\label{tab:conditions}
\centering
\scriptsize
\setlength{\tabcolsep}{1.2pt}
\begin{tabularx}{\columnwidth}{@{}lXccc@{}}
\toprule
Variant & Residual target
& \shortstack{Overall}
& \shortstack{Navigation}
& \shortstack{Port. manip.} \\
\midrule
Parent & None
& $3.86\!\pm\!0.06$
& $3.82\!\pm\!0.06$
& $3.99\!\pm\!0.19$ \\
Base-VO & Base (VO)
& $3.91\!\pm\!0.03$
& $3.90\!\pm\!0.07$
& $3.95\!\pm\!0.16$ \\
All-Action-VO & All actions (VO)
& $3.83\!\pm\!0.05$
& $3.82\!\pm\!0.11$
& \textbf{$3.88\!\pm\!0.22$} \\
VOMMI & Base (VO) + manip.
& \textbf{$3.79\!\pm\!0.02$}
& \textbf{$3.76\!\pm\!0.07$}
& $3.89\!\pm\!0.20$ \\
\bottomrule
\end{tabularx}
\vspace {-2.0em}
\end{table}

\begin{figure}[htb]
    \vspace {-1.0em}
    \centering
    \includegraphics[width=0.95\columnwidth]{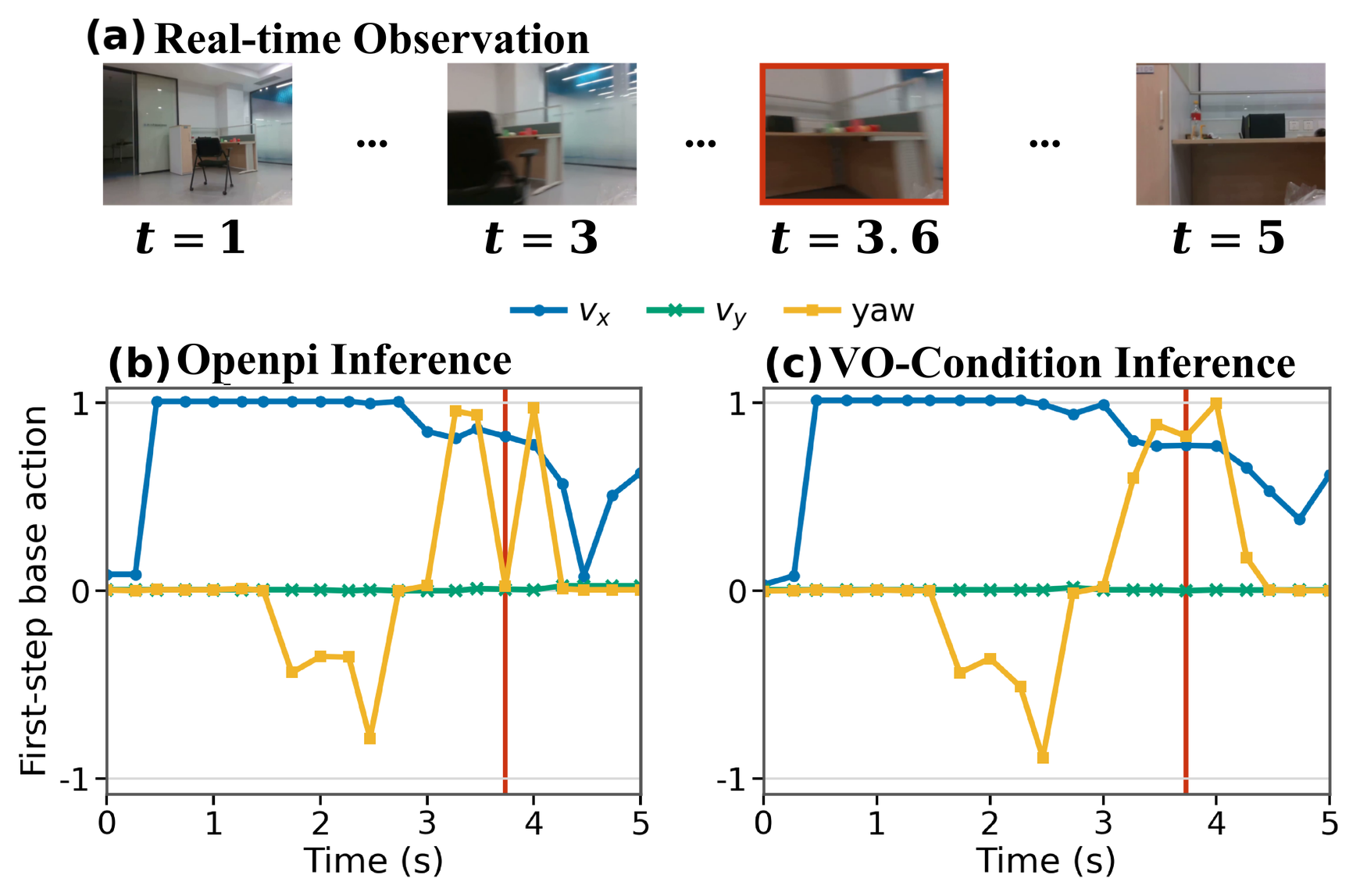}
    \vspace {-1.0em}
    \caption{Open-loop comparison using global observations and
    first-step base actions $(v_x,v_y,\mathrm{yaw})$, with the residual VO turn highlighted at
    $t=3.73$ s.}
    \label{fig:q3-openloop}
\end{figure}

\textbf{Local-action robustness (Q3).}
Figure~\ref{fig:q3-openloop} further examines how motion conditioning changes
the temporal action response on one real-robot episode. At the highlighted
observation, the OpenPI baseline has returned to near-zero yaw, whereas the
VO-conditioned policy retains a positive yaw command. This case study shows
that local-motion conditioning materially changes policy dynamics.


\textbf{VO conditioning and residual design (Q4).}
Table~\ref{tab:conditions} compares Parent (no VO or residual),
Base-VO (a VO-conditioned base residual), All-Action-VO
(VO-conditioned residuals for all actions), and VOMMI
(a VO-conditioned base residual plus a manipulation residual driven only by $h_{t,k}$).
We report overall, navigation, and portable-manipulation flow losses.
Within-source shuffled VO provides a paired motion-content check for the VO variants.

\section{Limitation and Conclusion}
Portable VOMMI demonstrations support local-motion prediction and produce actions that closely match the teleoperation references in in-domain real-world tasks, demonstrating effectiveness on long-horizon mobile manipulation tasks.
These benefits are most pronounced in settings with substantial navigation demands, while gains are relatively limited in tasks dominated by manipulation.
Our method has been validated on the M20S platform, while further cross-embodiment evaluation remains necessary.

VOMMI provides a portable RGB framework for learning long-horizon mobile
manipulation from robot-free demonstrations. It combines provenance-aware
masked observations, causal body-frame motion, planar Body-to-Hand
compensation, and action-group residuals that condition base actions on visual
motion. Collection-time tracking poses supervise the RGB motion estimator and
the local TCP $\SEthree$ predictor during data construction. Together, these
components form a unified pipeline from portable demonstrations to whole-body
policy learning.

\section*{Acknowledgments}
\textbf{AI-Use Disclosure.} OpenAI Codex assisted with drafting and revising selected manuscript text, refining figure captions, checking reference consistency, formatting, and generating plotting code from author-provided statistics. The authors reviewed and verified all AI-assisted content and take full responsibility for the final manuscript.

\bibliographystyle{IEEEtran}
\bibliography{references}
\end{document}